# Improved Performance of Unsupervised Method by Renovated K-Means

P.Ashok
Research Scholar,
Bharathiar University,
Coimbatore
Tamilnadu, India.
ashokcutee@gmail.com

Dr.G.M Kadhar Nawaz
Department of Computer Application
Sona College of Technology,
Salem, Tamilnadu, India
nawazse@yahoo.co.in

E.Elayaraja
Department of Computer Science
Periyar University, Salem-11
Tamilnadu, India
elayarajaphd.e@gmail.com

V.Vadivel
Department of Computer Science
Periyar University, Salem-11
Tamilnadu, India
v.vadivelmsc@gmail.com

**Abstract:** *Clustering is a separation of data into groups of similar objects. Every group called cluster consists of objects that are similar to one another and dissimilar to objects of other groups. In this paper, the K-Means algorithm is implemented by three distance functions and to identify the optimal distance function for clustering methods. The proposed K-Means algorithm is compared with K-Means, Static Weighted K-Means (SWK-Means) and Dynamic Weighted K-Means (DWK-Means) algorithm by using Davis Bouldin index, Execution Time and Iteration count methods. Experimental results show that the proposed K-Means algorithm performed better on Iris and Wine dataset when compared with other three clustering methods.*

## I. INTRODUCTION

### A. Clustering

Clustering [9] is a technique to group together a set of items having similar characteristics. Clustering can be considered the most important **unsupervised learning** problem. Like every other problem of this kind, it deals with finding a *structure* in collection of unlabeled data. A loose definition of clustering could be "the process of organizing objects into groups whose members are similar in some way". A *cluster* is therefore a collection of objects which are "similar" between them and are "dissimilar" to the objects belonging to other clusters. We can show this with a simple graphical example.

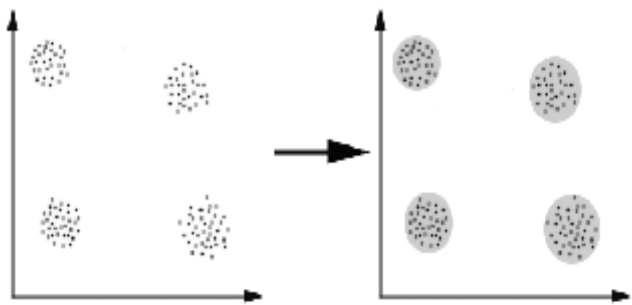

**Figure 1. cluster analysis**

In this case, we easily identify the 4 clusters into which the data can be divided. The similarity criterion is distance of two or more objects belong to the same cluster if they are "close" according to a given distance (in the case of geometrical distance). This is called the distance-based clustering. Another kind of clustering is conceptual clustering. Two or more objects belong to the same cluster if it defines a concept common to all that objects. In other words, objects are grouped according to their relevance to descriptive concepts, not according to simple similarity measures.

### B. Goals of Clustering

The goal of clustering is to determine the intrinsic grouping in a set of unlabeled data. The main requirements that a clustering algorithm should satisfy are:

- scalability
- Dealing with different types of attributes.
- Discovering clusters with arbitrary shape.
- Ability to deal with noise and outliers.
- Insensitivity to order of input records.
- High dimensionality and
- Interpretability and usability.

Clustering has number of problems. Few among them are listed below.

- Current clustering techniques do not address all the requirements adequately (and concurrently).
- Dealing with large number of dimensions and large number of data items can be problematic because of time complexity.
- The effectiveness of the method depends on the definition of "distance" (for distance-based clustering) and
- If an obvious distance measure doesn't exist we must "define" it, which is not always easy, especially in multi-dimensional spaces.

A large number of techniques have been proposed for forming clusters from distance matrices. Hierarchical techniques, optimization techniques and mixture model are the most important types. We discuss the first two types here. We will discuss mixture models in a separate note that includes their use in classification and regression as well as clustering.

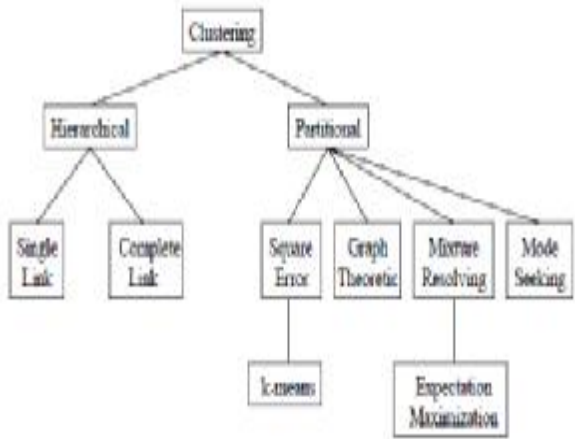

**Figure2. Taxonomy of Clustering Approaches**

At a high level, we can divide clustering algorithms into two broad classes as mentioned in the below section.

### C. Clustering Methods

#### 1) Hierarchical Clustering [3]

We begin assuming that each point is a cluster by itself. We repeatedly merge nearby clusters, using some measure of how close two clusters are (e.g., distance between their centroids), or how good a cluster the resulting group would be (e.g., the average distance of points in the cluster from the resulting centroids).

A hierarchical algorithm [8] yields a dendogram, representing the nested grouping of patterns and similarity levels at which groupings change. The dendogram can be broken at different levels to yield different clustering of the data. Most hierarchical clustering algorithms are variants of the single-link, complete-link, and minimum-variance algorithms. The single-link and complete link algorithms are most popular. These two algorithms differ in the way they characterize the similarity between a pair of clusters. In the single-link method, the distance between two clusters is the minimum of the distances between all pairs of patterns drawn from the two clusters (one pattern from the first cluster, the other from the second). In the complete-link algorithm, the distance between two clusters is the minimum of all pair wise distances between patterns in the two clusters. In either case, two clusters are merged to form a larger cluster based on minimum distance criteria. The complete-link algorithm produces tightly bound or compact clusters. The single-link algorithm, by contrast, suffers from a chaining effect. It has a tendency to produce clusters that are straggly or elongated. The clusters obtained by the complete link algorithm are more compact than those obtained by the single-link algorithm.

#### 2) Partitional Clustering

A Partitional clustering [9] algorithm obtains a single partition of the data instead of a clustering structure, such as dendogram produced by a hierarchical technique. Partitional methods have advantages in applications involving large data sets for which the construction of a dendogram is computationally prohibitive. A problem accompanying the use of a Partitional algorithm is the choice of the number of desired output clusters. The Partitional technique usually produce clusters by optimizing a criterion function defined either locally (on a subset of the patterns) or globally (defined over all of the patterns). Combinatorial search of the set of possible labeling for an optimum value of a criterion is clearly computationally prohibitive. In practice, therefore, the algorithm is typically run multiple times with different starting states, and the best configuration obtained from all of the runs issued as the output clustering.

### D. Applications of Clustering

Clustering algorithms can be applied in many fields, such as the ones that are described below

- Marketing: Finding groups of customers with similar behavior gives a large database of customer data containing their properties and past buying records.
- Biology: Classification of plants and animals given their features given.
- Libraries: Book ordering
- Insurance: Identifying groups of motor insurance policy holders with a high average claim cost frauds.
- City-planning: Identify groups of houses according to their house type, value and geographical location.
- Earthquake studies: Clustering observed earthquake epicenters to identify dangerous zones and
- WWW: Document classification, clustering weblog data to discover groups of similar access patterns.

The rest of this paper is organized as follows: section II describes three different Distance functions, Execution Time method and the clustering algorithms viz. for K-Means, Weighted K-Means and Proposed K-Means Clustering algorithms have been studied and implemented. Section III presents the experimental analysis conducted on various data sets of UCI data repository and section IV concludes the paper.

## II. Partitional Clustering Algorithms

### A. K-Means Algorithm

The K-Means [2] algorithm is an iterative procedure for clustering which requires an initial classification of data. It computes the center of each cluster, and then computes new partitions by assigning every object to the cluster whose center is the closest to that object. This cycle is repeated during a given number of iterations or until the assignment has not changed during one iteration. This algorithm is based on an approach where a random set of cluster base is selected from the original dataset, and each element update the nearest element of the base with the average of its attributes. The K-Means is possibly the most commonly-used clustering algorithm. It is most effective for relatively smaller data sets. The K-Means finds a locally optimal solution by minimizing a distance measure between each data and its nearest cluster center. The basic K-Means algorithm is commonly measured by any of intra-cluster or inter-cluster criterion. A typical intra-cluster criterion is the squared-error criterion. It is the most commonly used and a good measure of the within-cluster variation across all the partitions. The process iterates through the following steps:

- Assignment of data to representative centers upon minimum distance and
- Computation of the new cluster centers.

K-Means clustering is computationally efficient for large data sets with both numeric and categorical [3] attributes.

## 1) Working Principle

The K-Means [9] algorithm works as follows. First, it iteratively selects k of the objects, each of which initially represents a cluster mean or center. For each of the remaining objects, an object is assigned to the cluster to which it is the most similar, based on the distance between the object and the cluster mean. It then computes the new mean for each cluster. This process iterates until the criterion function converges. Typically, the Euclidean distance is used.

---

**Algorithm 1: K-Means**

1. Initialize the number of clusters k.
2. Randomly selecting the centroids in the given dataset ($c_1, c_2, ..., c_k$).
3. Compute the distance between the centroids and objects using the Euclidean Distance equation.

$$d_{ij} = \|x_i - c_k\|^2$$

4. Update the centroids.
5. Stop the process when the new centroids are nearer to old one. Otherwise, go to step-3.

---

### B. Weighted K-Means Clustering Algorithm

Weighted K-Means [9] algorithm is one of the clustering algorithms, based on the K-Means algorithm calculating with weights. This algorithm is same as normal K-Means algorithm just adding with weights. Weighted K-Means attempts to decompose a set of objects into a set of disjoint clusters taking into consideration the fact that the numerical attributes of objects in the set often do not come from independent identical normal distribution. Weighted K-Means algorithms are iterative and use hill-climbing to find an optimal solution (clustering) and thus usually gives converge to a local minimum.

First, calculate the weights for the corresponding centroids in the data set and then calculate the distance between the object and the centroid with the weights of the centroids. This method is called the Weighted K-Means algorithm.

In the Weighted K-Means algorithm the weights can be classified into two types as follows.

- **Dynamic Weights:** In the dynamic weights, the weights are determined during the execution time which can be changed at runtime.
- **Static Weights:** In the static weights, the weights are not changed during the runtime.

## 2) Working Principle

The Weighted K-Means [9] algorithm works as defined below. First, it iteratively selects K of the objects, each of which initially represents a cluster mean or center. In the selecting centroids we calculate the weights using the Weighted K-Means algorithm. For each of the remaining objects, an object is assigned to the cluster to which it is the most similar, based on the distance between objects and centroids with weights of the corresponding object. It then computes the new mean for each cluster. This process iterates until the criterion function converge. Typically, the Euclidean distance is used in the clustering process. In the Euclidean distance, we can calculate the Dynamic weights based on the particular centroids.

Calculate the Sum $s_i = \sum_{j=1}^{n} c_{ij}$, $i = 1, 2, ..., k$ and $w_i = (s_i - c_{ij})/s_i$ where j=1, 2, …, n.

$w^{(i)}$ is corresponding weight vector to the $c^{(i)}$

$$d_{ij} = \|w_i * (x_i - c_k)\|^2$$

Using this equation we can calculate the distance between the centroids and the weights.

The weighted K-Means clustering algorithms are explained in the following section. In the static weighted K-Means, the weight is fixed 1.5 as constant but the dynamic weight is calculated by the above equation and the weighted K-Means clustering algorithm is explained in algorithm 2.

---

**Algorithm 2: Weighted K-Means**

Steps:

1. Initialize the number of clusters k.
2. Randomly selecting the centroids ($c_1, c_2, ..., c_k$) in the data set.
3. Calculating the weights $w^{(i)}$ of the corresponding centroids ($c_1, c_2, ..., c_k$).
4. Calculate Sum $s_i = \sum_{j=1}^{n} s_i$, $i = 1, 2, ..., k$ and $w_i = (s_i - c_{ij})/s_i$ where j=1, 2… n. Where $w_i$ is corresponding weight vector to the $c_i$.
5. Find the distance between the centroids using the Euclidean Distance equation
$$d_{ij} = \|w_i * (x_i - c_k)\|^2$$
6. Update the centroids Stop the process when the new centroids are nearer to old one. Otherwise, go to step-4.

---

### C. Proposed K-Means

In the proposed method, first, it determines the initial cluster centroids by using the equation which is given in the following algorithm 3. The Proposed K-Means algorithm is improved by selecting the initial centroids manually instead of selecting centroids by randomly. It selects 'K' objects and each of which initially represents a cluster mean or centroids. For each of the remaining objects, an object is assigned to the cluster to which it is the most similar based on the distance between the object and the cluster mean. It then computes the new mean for each cluster. This process iterates until the criterion function converges. In this paper the Proposed K-Means algorithm is implemented instead of traditional K-Means as explained in the algorithm 3.

---

**Algorithm 3: Proposed K-Means**

Steps:

1. Using Euclidean distance as a dissimilarity measure, compute the distance between every pair of all objects as follow.

$$d_{ij} = \sqrt{\sum_{a=1}^{p}(X_{ia} - X_{ja})^2} \quad i\&j = 1,\ldots,n; \quad (1)$$

2. Calculate **M**$_{ij}$ to make an initial guess at the centres of the clusters

$$M_{ij} = \frac{d_{ij}}{\sum_{i=1}^{n} d_{ij}} \quad i\&j = 1,\ldots,n. \quad (2)$$

3. Calculate $\sum_{i=1}^{n} M_{ij}^2 \quad (j=1\ldots n)$ (3) at each **object** and sort them in ascending order.
4. Select K objects having the minimum **value** as initial cluster centroids which are determined by the above equation. Arbitrarily choose $k$ data points from $D$ as initial centroids.
5. Assign each point $d_i$ to the cluster which has the closest centroid.
6. Calculate the new mean for each cluster.
7. **Repeat step 5** and **step 6 until convergence** criteria is met.

D. **Cluster validity Measure**

Many criteria have been developed for determining cluster validity all of which have a common goal to find the clustering which results in compact clusters which are well separated. Clustering validity is a concept that is used to evaluate the quality of clustering results. The clustering validity index may also be used to find the optimal number of clusters and measure the compactness and separation of clusters.

1) **Davies-Bouldin Index**

In this paper, **DAVIS BOULDIN** index [1 and 6] has been chosen as the cluster validity measure because it has been shown to be able to detect the correct number of clusters in several experiments. Davis Bouldin validity is the combination of two functions. First, calculates the compactness of data in the same cluster and the second, computes the separateness of data in different clusters.

This index (Davies and Bouldin, 1979) is a function of the ratio of the sum of within-cluster scatter to between-cluster separation. If $dp_i$ is the dispersion of the cluster Pi, and $dv_{ij}$ denotes the dissimilarity between two clusters $P_i$ and $P_j$, then a cluster similarity matrix FR = { FR$_{ij}$ , (i, j) = 1. 2 …..C} is defined as:

$$FR_{ij} = \frac{dp_i + dp_j}{dv_{ij}}$$

The dispersion dpi can be seen as a measure of the radius of $P_i$,

$$dp_i = \left(\frac{1}{n_i}\sum_{x \in p_i} \|x - V_i\|^2\right)^{\frac{1}{2}}$$

Where $n_i$ is the number of objects in the i$^{th}$ cluster. $V_i$ is the centroid of the i$^{th}$ cluster. dv$_{ij}$ describes the dissimilarity between $P_i$ and $P_j$,

$$dv_{ij} = \|V_i - V_j\|^2$$

The corresponding DB index is defined as:

$$DB_{FR} = \frac{1}{c}\sum_{i=1}^{c} FR_i$$

c is the number of cluster. Hence the ratio is small if the clusters are compact and far from each other. Consequently Davies-Bouldin index will have a small value for a good clustering.

E. **Distance Measures**

Many clustering methods use distance measures [7] to determine the similarity or dissimilarity between any pair of objects. It is useful to denote the distance between two instances $x_i$ and $x_j$ as: **d ($x_i$, $x_j$)**. A valid distance measure should be symmetric and obtains its minimum value (usually zero) in case of identical vectors. The distance functions are classified into 3 types.

1) **Manhattan Distance**

The Minkowski distance or the **L1 distance** is called the Manhattan distance [5] and is described in the below equation.

$$d_i(X,Y) = \sum_{i=1}^{n} |x_i - y_i|$$

It is also known as the **City Block distance**. This metric assumes that in going from one pixel to the other, it is only possible to travel directly along pixel grid lines. Diagonal moves are not allowed.

2) **Euclidian Distance[5]**

This is the most familiar distance that we use. To find the shortest distance between two points $(x_1, y_1)$ and $(x_2, y_2)$ in a two dimensional space that is

$$d_i(X,Y) = \sqrt{\sum_{i=1}^{n} (x_i - y_i)^2}$$

3) **Chebyshev Distance**

The Chebyshev [10] distance between two vectors or points $p$ and $q$, with standard coordinates p$_i$ and q$_i$, respectively.

$$D_{Chebyshev}(p,q) = max_i(|x_i - y_i|)$$

III. **EXPERIMENTAL ANALYSIS AND DISCUSSION**

This section describes, the data sets used to analyze the methods studied in sections II and III, which are arranged in the form of a list in Table1.

A. **Datasets**
   1) **Iris**

The iris dataset contains the information about the iris flower. The data set contains 150 samples with four attributes. The dataset is collected from the location which is given in the link. **http://archive.ics.uci.edu/ml/ machine-learning-databases /iris/ iris.data**

2) **Ecoli**

The Ecoli dataset contains protein localization sites having 350 samples with 8 attributes. The dataset is collected from the location which is given in the link. **http://archive.ics.uci.edu/ml/machine-learning-databases /ecoli/ecoli.data**

3) **Yeast**

The yeast dataset contains 1400 samples with 8 attributes. The dataset is collected from the location which is given in the link. **http://archive.ics.uci.edu/ml/machine-learning-databases /yeast/yeast.data**

4) **Wine**

The dataset contains the information about to determine the origin of wines. It contains 200 samples with 13 attributes and the dataset is collected from the location which is given in the link. **http://archive.ics.uci.edu/ml/machine-learning-databases/ yeast/wine.data**

### B. Comparative Study and Performance analysis

The four clustering algorithms are K-Means, Static Weighted K-Means **(SWK-Means),** Dynamic Weighted K-Means **(DWK-Means)** and **Proposed K-Means** that are used to clustering the data sets. To access the quality of the clusters, the Davis Bouldin measure has been used.

1) **Performance of Distance functions**

The K-Means clustering method is executed by three different distance functions are Manhattan, Euclidean and chebyshev with iris dataset are used and select the centroid value (K) from 2 to 10. The obtained results are listed in the table I given below.

**TABLE I.    Distance Functions**

| S.No | Clusters | Distance Functions | | |
|---|---|---|---|---|
| | | *Manhattan* | *Euclidean* | *Chebyshev* |
| 1 | 2 | 0.510 | 0.524 | 0.658 |
| 2 | 3 | 0.653 | 0.416 | 0.548 |
| 3 | 4 | 0.758 | 0.634 | 0.811 |
| 4 | 5 | 0.912 | 0.589 | 0.987 |
| 5 | 6 | 0.933 | 0.712 | 1.023 |
| 6 | 7 | 0.847 | 0.689 | 0.956 |
| 7 | 8 | 0.935 | 0.881 | 1.095 |
| 8 | 9 | 0.874 | 0.643 | 0.912 |
| 9 | 10 | 0.901 | 0.773 | 0.845 |

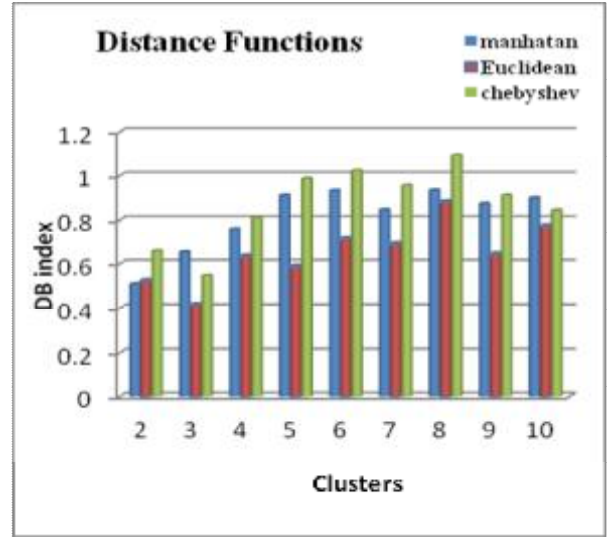

**Figure4. Distance functions analysis chart for K-Means**

From the figure 4 shows that, the various distance functions for K-Means are studied and compared with the dataset "Ecoli". The K-Means clustering methods are executed with varying cluster centroids K from 2 to 10 with the data set. It clearly shows that the Euclidean distance function obtained the minimum index values for most of the different clusters values. Hence the Euclidean distance function is better for clustering algorithms than other distance functions.

2) **Performance of Clustering Algorithms**

The four algorithms are studied in the section II which are implemented by the software **MATLAB 2012 (a).** All the methods are executed and compared by Ecoli, Iris, Yeast and Wine dataset. The Davis Bouldin index is used to determine the performance of the clustering algorithms. The results are obtained from the various clustering algorithms and are listed in the Table II below.

**TABLE II.    Davis Bouldin index analysis**

| S.No | Data set | Davis Bouldin Index | | | |
|---|---|---|---|---|---|
| | | *K-Means* | *Static Weight K-Means* | *Dynamic Weight K-Means* | *Proposed K-Means* |
| 1 | Ecoli | 0.71 | 0.95 | 0.84 | 0.65 |
| 2 | Iris | 0.65 | 0.49 | 0.53 | 0.73 |
| 3 | Yeast | 0.86 | 0.79 | 0.84 | 0.66 |
| 4 | Wine | 0.55 | 0.74 | 0.68 | 0.45 |

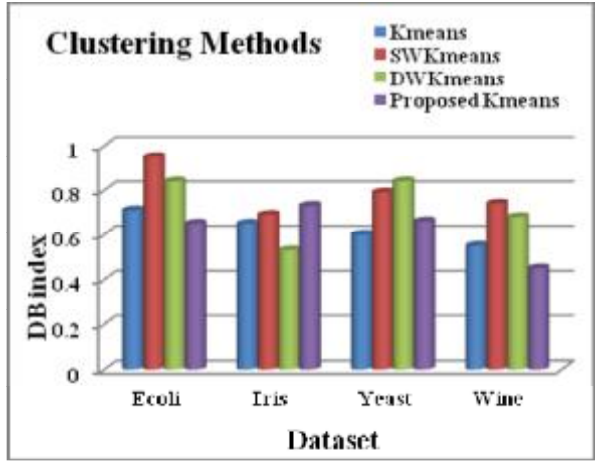

Figure3. Clustering methods chart for various dataset

From the figure3 shows that, the four clustering algorithms are executed by the four different data set called **Ecoli, Iris, Yeast, Wine** with the constant Cluster Centroid (K) whose value is 5. The proposed K-Means algorithm obtained the minimum DB index values for the Ecoli, Wine data set also obtained next minimum index which is more than that of other algorithm. Hence the proposed K-Means clustering algorithm obtained good clustering results.

#### 3) Execution Time Measure

Different Clustering algorithms are compared for their performances using the time required to cluster the dataset. The execution time is varying while selecting the number of initial cluster centroids. The execution time is increased when the number of cluster centroid is increased. The obtained results are depicted in the following Table III

TABLE III. Execution Time

| S.No | Cluster | Execution Time (in sec) | | | |
|---|---|---|---|---|---|
| | | K-Means | Static Weight K-Means | Dynamic Weight K-Means | Refined K-Means |
| 1 | 3 | 1.54 | 1.66 | 1.45 | 1.31 |
| 2 | 6 | 1.95 | 2.87 | 2.28 | 2.05 |
| 3 | 9 | 3.78 | 4.15 | 3.94 | 3.45 |
| 4 | 12 | 4.45 | 5.12 | 4.18 | 3.94 |
| 5 | 15 | 5.7 | 6.45 | 5.48 | 4.71 |

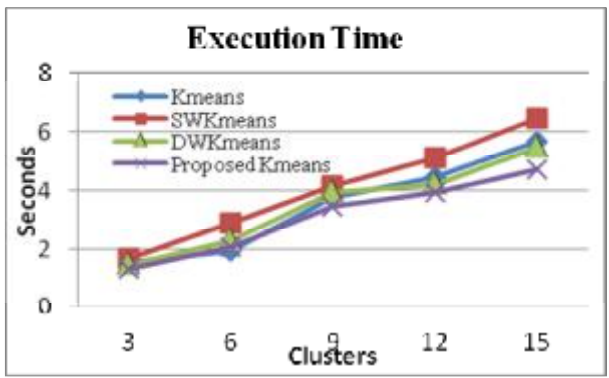

Figure5. Execution Time chart for clustering algorithms

From the figure5 shows that, the performance of the four clustering algorithms are executed by the Iris dataset with varying the cluster centroids from 3 to 15. The proposed K-Means algorithm obtained the minimum execution time for most of the clustering centroids. The SWK-Means clustering algorithm does not produce minimum execution time for all various K values, but the other two algorithms produce minimum execution time for some K centroids values. Hence the proposed K-Means clustering executed in the minimum execution time and performed better than other 3 algorithms.

#### 4) Iteration Count Analysis

The four Clustering algorithms are compared for their performance using Iteration count method with the Wine dataset. The Iteration count is defined as that the number of times the clustering algorithm is executed until the convergence criteria is met. The cluster centres are increased each time by 3 and the number of iterations for each clustering algorithms are obtained and listed in the below Table IV.

TABLE IV. Iteration count Analysis

| S. No | Cluster | Iteration Level | | | |
|---|---|---|---|---|---|
| | | K-Means | Static Weight K-Means | Dynamic Weight K-Means | Proposed K-Means |
| 1 | 2 | 8 | 7 | 11 | 6 |
| 2 | 3 | 6 | 8 | 5 | 7 |
| 3 | 4 | 14 | 18 | 15 | 9 |
| 4 | 5 | 9 | 13 | 11 | 8 |
| 5 | 6 | 4 | 8 | 11 | 6 |
| 6 | 7 | 15 | 13 | 17 | 11 |
| 7 | 8 | 13 | 10 | 15 | 6 |
| 8 | 9 | 7 | 6 | 11 | 10 |
| 9 | 10 | 14 | 17 | 14 | 8 |

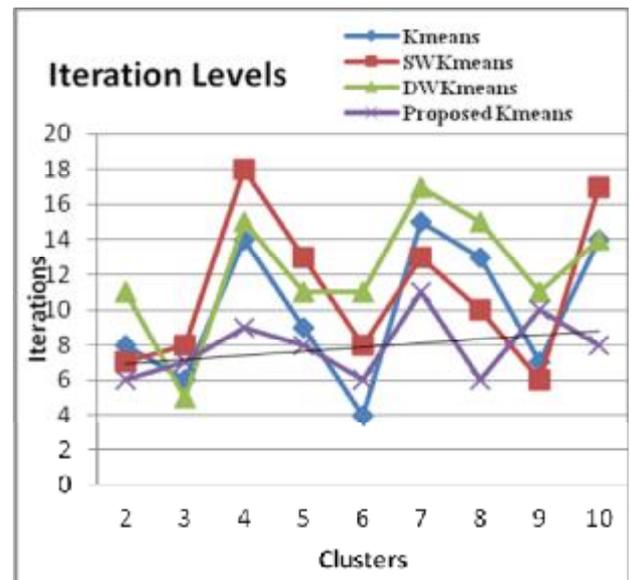

Figure6. Iteration levels chart for Clustering Algorithms

From the figure6, the iteration levels are identified by the four algorithms by setting the cluster centroid from 2 to 10 with the yeast dataset. The K-Means and proposed K-Means clustering methods performed well in minimum iterations.

The Proposed K-Means method performed better than K-Means for most of the Clustering centroids (K) values.

## IV. CONCLUSION

In this paper, the four different Clustering methods in the Partitional clustering are studied by applying the K-Means algorithm with three different distance functions to find the optimal distance function for clustering process. One of the demerits of K-Means algorithm is random selection of initial centroids of desired clusters. This was overcome by proposed K-Means with initial cluster centroid selection process for finding the initial centroids to avoid the selecting centroids randomly and it produces distinct better results. The four clustering algorithms are executed with four different dataset but the Proposed K-Means method performs very well and obtains minimum DB index value. The Execution time and iteration count is compared with the four different clustering algorithms and different cluster values. The Proposed K-Means achieved less execution time and minimum iteration count than K-Means, Static Weighted K-Means (**SWK-Means**), and Dynamic Weighted K-Means (**DWK-Means**) clustering methods. Therefore, the proposed K-Means clustering method can be applied in the application area and various cluster validity measures can be used to improve the cluster performance in our future work.

## REFERENCES

[1] Davies & Bouldin, 1979. Davies, D.L., Bouldin, D.W., (2000) "A cluster separation measure." IEEE Trans.Pattern Anal. Machine Intell., 1(4), 224-227.

[2] HARTIGAN, J. and WONG, M. 1979. Algorithm AS136: "A k-means clustering algorithm". Applied Statistics, 28, 100-108.

[3] HEER, J. and CHI, E. 2001. "Identification of Web user traffic composition using multimodal clustering and information scent.", 1st SIAM ICDM, Workshop on Web Mining, 51-58, Chicago, IL.

*[4]* JAIN A.K, MURTY M.N. And FLYNN P.J. *Data Clustering: A Review ACM Computing Surveys*, *Vol. 31*, *No. 3*, *September 1999.*

[5] Maleq Khan - Fast Distance Metric Based Data Mining Techniques Using P-trees: k-Nearest-Neighbor Classification and k-Clustering.

[6] *Maria Halkidi*, Yannis Batistakis, and Michalis Varzirgiannis, '*On clustering validation techniques*', *Journal of Intelligent Information Systems*, *17(2-3)*, *107–145*, (*2001*).

[7] Rokach L O Maimon - Data mining and knowledge discovery handbook, 2005 – Springer.

[8] Sauravjoyti Sarmah and Dhruba K. Bhattacharyya. May 2010 "An Effective Technique for Clustering Incremental Gene Expression data", IJCSI International Journal of Computer Science Issues, Vol. 7, Issue 3, No. 3.

[9] Thangadurai, K. e.a. 2010. A Study On Rough Clustering. In «Global Journal of Computer Science and Technology», Vol. 10, Issue 5.

[10] http://en.wikipedia.org/wiki/Chebyshev_distance